\documentclass[11pt,a4paper]{article}
\usepackage[hyperref]{emnlp-ijcnlp-2019}
\usepackage{times}
\usepackage{latexsym}

\usepackage{url}

\usepackage{graphicx}
\usepackage{amsfonts}
\usepackage{amsmath}
\usepackage{amssymb}
\usepackage{pifont}
\usepackage{nicefrac}
\usepackage{multirow}
\usepackage{subfigure}
\usepackage[font=small]{caption}
\usepackage{todonotes}
\usepackage{makecell}

\usepackage{enumitem}
\setenumerate[1]{itemsep=0.30pt,partopsep=0pt,parsep=\parskip,topsep=5pt}
\setitemize[1]{itemsep=0.30pt,partopsep=00pt,parsep=\parskip,topsep=5pt}
\setdescription{itemsep=0.30pt,partopsep=0pt,parsep=\parskip,topsep=5pt}

\usepackage{array}
\newcolumntype{H}{>{\setbox0=\hbox\bgroup}c<{\egroup}@{}}

\aclfinalcopy % Uncomment this line for the final submission

\title{Improving Deep Transformer\\ with Depth-Scaled Initialization and Merged Attention}

\author{Biao Zhang$^1$ \quad Ivan Titov$^{1,2}$ \quad Rico Sennrich$^{3,1}$ \bigskip\\
  $^1$School of Informatics, University of Edinburgh \\
  $^2$ILLC, University of Amsterdam \\
  $^3$Institute of Computational Linguistics, University of Zurich \\
  \texttt{B.Zhang@ed.ac.uk, ititov@inf.ed.ac.uk, sennrich@cl.uzh.ch}
  }

\date{}

\begin{document}
\maketitle
\begin{abstract}

The general trend in NLP is towards increasing model capacity and performance via deeper neural networks.
However, simply stacking more layers of the popular Transformer architecture for machine translation results in poor convergence and high computational overhead. Our empirical analysis suggests that convergence is poor due to \textit{gradient vanishing} caused by the interaction between residual connections and layer normalization. We propose depth-scaled initialization (DS-Init), which decreases parameter variance at the initialization stage, and reduces output variance of residual connections so as to ease gradient back-propagation through normalization layers. To address computational cost, we propose a merged attention sublayer (MAtt) which combines a simplified average-based self-attention sublayer and the encoder-decoder attention sublayer on the decoder side. Results on WMT and IWSLT translation tasks with five translation directions show that deep Transformers with DS-Init and MAtt can substantially outperform their base counterpart in terms of BLEU (+1.1 BLEU on average for 12-layer models), while matching the decoding speed of the baseline model thanks to the efficiency improvements of MAtt.\footnote{Source code for reproduction is available at \url{https://github.com/bzhangGo/zero}}

\end{abstract}

\section{Introduction}

\begin{figure}[t]
\centering
\includegraphics[scale=0.55]{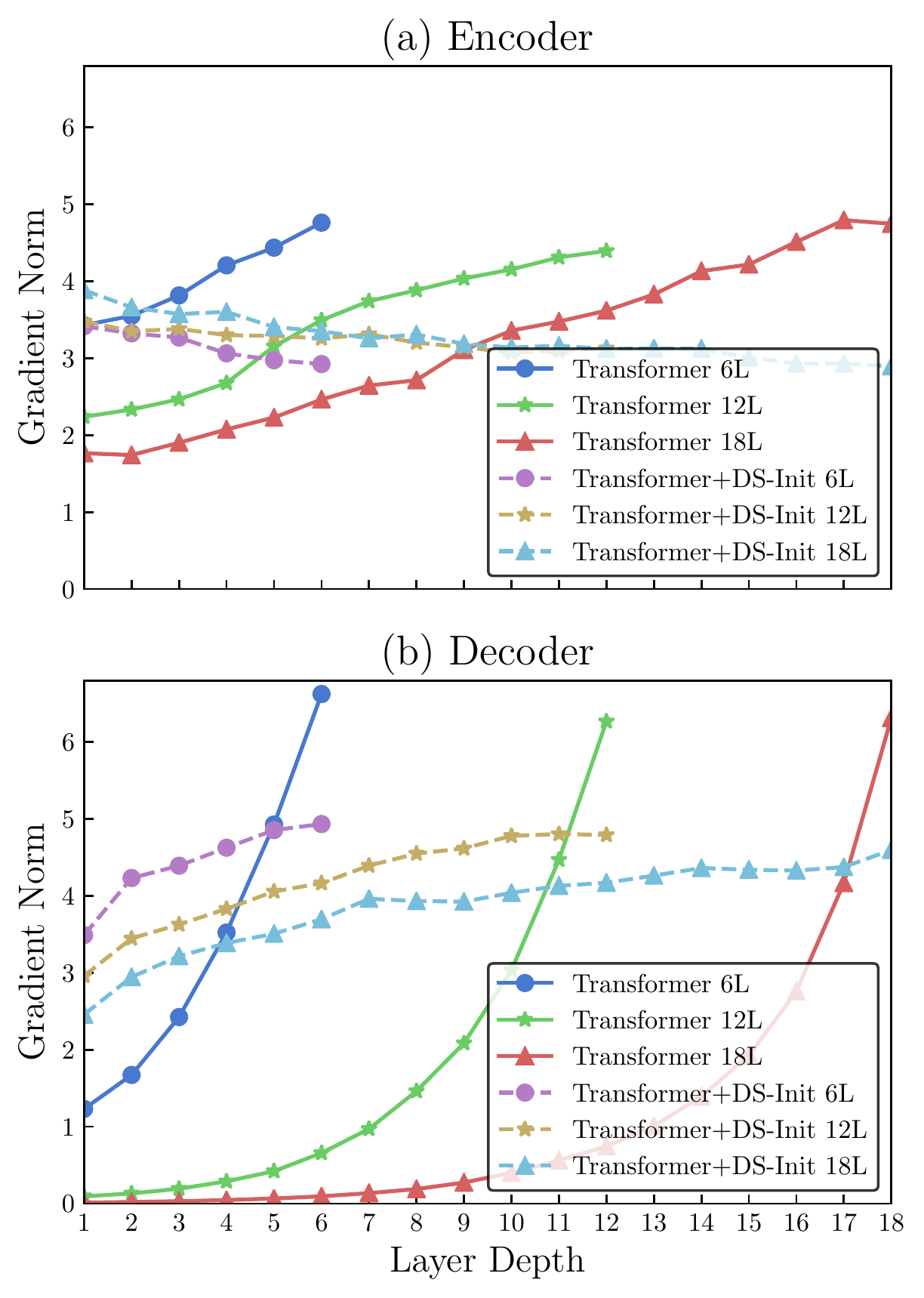}
\caption{\label{fig_encdec_gradnorm} Gradient norm (y-axis) of each encoder layer (top) and decoder layer (bottom) in Transformer with respect to layer depth (x-axis). Gradients are estimated with $\sim$3k target tokens at the beginning of training. ``\textit{DS-Init}'': the proposed depth-scaled initialization. ``\textit{6L}'': 6 layers. Solid lines indicate the vanilla Transformer, and dashed lines denote our proposed method. During back-propagation, gradients in Transformer gradually vanish from high layers to low layers.}
\end{figure}

The capability of deep neural models of handling complex dependencies has benefited various artificial intelligence tasks, such as image recognition where test error was reduced by scaling VGG nets~\cite{Simonyan15} up to hundreds of convolutional layers~\cite{DBLP:journals/corr/HeZRS15}. In NLP, deep self-attention networks have enabled large-scale pretrained language models such as BERT~\cite{devlin2018bert} and GPT~\cite{radford2018improving} to boost state-of-the-art (SOTA) performance on downstream applications. By contrast, though neural machine translation (NMT) gained encouraging improvement when shifting from a shallow architecture~\cite{bahdanau+al-2014-nmt} to deeper ones ~\cite{zhou-etal-2016-deep,45610,8493282,P18-1008}, the Transformer~\cite{NIPS2017_7181}, a currently SOTA architecture, achieves best results with merely 6 encoder and decoder layers, and no gains were reported by \citet{NIPS2017_7181} from further increasing its depth on standard datasets.

We start by analysing why the Transformer does not scale well to larger model depth.
We find that the architecture suffers from gradient vanishing as shown in Figure \ref{fig_encdec_gradnorm}, leading to poor convergence. An in-depth analysis reveals that the Transformer is not norm-preserving due to the involvement of and the interaction between residual connection (RC)~\cite{DBLP:journals/corr/HeZRS15} and layer normalization (LN)~\cite{lei2016layer}.

To address this issue, we propose depth-scaled initialization (DS-Init) to improve norm preservation. We ascribe the gradient vanishing to the large output variance of RC and resort to strategies that could reduce it without model structure adjustment. Concretely, DS-Init scales down the variance of parameters in the $l$-th layer with a discount factor of $\frac{1}{\sqrt{l}}$ at the initialization stage alone, where $l$ denotes the layer depth starting from 1. The intuition is that parameters with small variance in upper layers would narrow the output variance of corresponding RCs, improving norm preservation as shown by the dashed lines in Figure \ref{fig_encdec_gradnorm}. In this way, DS-Init enables the convergence of deep Transformer models to satisfactory local optima.

Another bottleneck for deep Transformers is the increase in computational cost for both training and decoding. To combat this, we propose a merged attention network (MAtt). MAtt simplifies the decoder by replacing the separate self-attention and encoder-decoder attention sublayers with a new sublayer that combines an efficient variant of average-based self-attention
 (AAN)~\cite{zhang-etal-2018-accelerating} and the encoder-decoder attention. We simplify the AAN by reducing the number of linear transformations, reducing both the number of model parameters and computational cost.
The merged sublayer benefits from parallel calculation of (average-based) self-attention and encoder-decoder attention, and reduces the depth of each decoder block.

We conduct extensive experiments on WMT and IWSLT translation tasks, covering five translation tasks with varying data conditions and translation directions.
Our results show that deep Transformers with DS-Init and MAtt can substantially outperform their base counterpart in terms of BLEU (+1.1 BLEU on average for 12-layer models), while matching the decoding speed of the baseline model thanks to the efficiency improvements of MAtt.

Our contributions are summarized as follows:
\begin{itemize}
    \item We analyze the vanishing gradient issue in the Transformer, and identify the interaction of residual connections and layer normalization as its source.
    \item To address this problem, we introduce depth-scaled initialization (DS-Init).
    \item To reduce the computational cost of training deep Transformers, we introduce a merged attention model (MAtt). MAtt combines a simplified average-attention model and the encoder-decoder attention into a single sublayer, allowing for parallel computation.
    \item We conduct extensive experiments and verify that deep Transformers with DS-Init and MAtt improve translation quality while preserving decoding efficiency.
\end{itemize}

\section{Related Work}

Our work aims at improving translation quality by increasing model depth. Compared with the single-layer NMT system~\cite{bahdanau+al-2014-nmt}, deep NMT models are typically more capable of handling complex language variations and translation relationships via stacking multiple encoder and decoder layers~\cite{zhou-etal-2016-deep,45610,britz-etal-2017-massive,P18-1008}, and/or multiple attention layers~\cite{8493282}. One common problem for the training of deep neural models are vanishing or exploding gradients. Existing methods mainly focus on developing novel network architectures so as to stabilize gradient back-propagation, such as the fast-forward connection~\cite{zhou-etal-2016-deep}, the linear associative unit~\cite{wang-etal-2017-deep}, or gated recurrent network variants \cite{Hochreiter:1997:LSM:1246443.1246450,Gers:00icannga,cho-EtAl:2014:EMNLP2014,di2018deep}. In contrast to the above recurrent network based NMT models, recent work focuses on feed-forward alternatives with more smooth gradient flow, such as convolutional networks~\cite{pmlr-v70-gehring17a} and self-attention networks~\cite{NIPS2017_7181}. 

The Transformer represents the current SOTA in NMT. It heavily relies on the combination of residual connections~\cite{DBLP:journals/corr/HeZRS15} and layer normalization~\cite{lei2016layer} for convergence. Nevertheless, simply extending this model with more layers results in gradient vanishing due to the interaction of RC and LN (see Section \ref{sec_vanish}). Recent work has proposed methods to train deeper Transformer models, including a rescheduling of RC and LN~\cite{vaswani2018tensor2tensor}, the transparent attention model~\cite{bapna-etal-2018-training} and the stochastic residual connection~\cite{Pham2019VeryDS}.
In contrast to these work, we identify the large output variance of RC as the source of gradient vanishing, and employ scaled initialization to mitigate it without any structure adjustment. The effect of careful initialization on boosting convergence was also investigated and verified in previous work~\cite{zhang2018residual,Child2019GeneratingLS,devlin2018bert,radford2018improving}.

The merged attention network falls into the category of simplifying the Transformer so as to shorten training and/or decoding time. Methods to improve the Transformer's running efficiency range from algorithmic improvements~\cite{Junczys-Dowmunt-wnmtmarian}, non-autoregressive translation~\cite{Gu2018NonAutoregressiveNM,Ghazvininejad2019ConstantTimeMT} to decoding dependency reduction such as average attention network~\cite{zhang-etal-2018-accelerating} and blockwise parallel decoding~\cite{stern2018blockwise}.
Our MAtt builds upon the AAN model, further simplifying the model by reducing the number of linear transformations, and combining it with the encoder-decoder attention. 
In work concurrent to ours, \citet{so2019evolved} propose the evolved Transformer which, based on automatic architecture search, also discovered a parallel structure of self-attention and encoder-decoder attention.

\section{Background: Transformer}

Given a source sequence $\mathbf{X}=\{x_1, x_2, \ldots, x_n\} \in \mathbb{R}^{n\times d}$,
 the Transformer predicts a target sequence $\mathbf{Y}=\{y_1, y_2, \ldots, y_m\}$ under the encoder-decoder framework. Both the encoder and the decoder in the Transformer are composed of attention networks, functioning as follows:
\begin{equation}\label{eq_att}
\begin{split}
    \textsc{Att}(\mathbf{Z}_x, \mathbf{Z}_y) & = \left[\text{softmax}(\frac{\mathbf{Q}\mathbf{K}^T}{\sqrt{d}}) \mathbf{V}\right] \mathbf{W}_o \\
    \mathbf{Q}, \mathbf{K}, \mathbf{V} & = \mathbf{Z}_x \mathbf{W}_q, \mathbf{Z}_y \mathbf{W}_k, \mathbf{Z}_y \mathbf{W}_v,
\end{split}
\end{equation}
where $\mathbf{Z}_x \in \mathbb{R}^{I\times d}$ and $\mathbf{Z}_y \in \mathbb{R}^{J\times d}$ are input sequence representations of length $I$ and $J$ respectively, $\mathbf{W}_* \in \mathbb{R}^{d\times d}$ denote weight parameters. The attention network can be further enhanced with multi-head attention~\cite{NIPS2017_7181}.

Formally, the encoder stacks $L$ identical layers, each including a self-attention sublayer (Eq.~\ref{eq_enc_self_att}) and a point-wise feed-forward sublayer (Eq.~\ref{eq_enc_ffn}): 
\begin{align}
\mathbf{\bar{H}}^l & = \textsc{Ln}\left(\textsc{Rc}\left(\mathbf{H}^{l-1}, \textsc{Att}(\mathbf{H}^{l-1}, \mathbf{H}^{l-1})\right)\right) \label{eq_enc_self_att}\\
\mathbf{H}^l &= \textsc{Ln}\left(\textsc{Rc}\left(\mathbf{\bar{H}}^l, \textsc{Ffn}(\mathbf{\bar{H}}^l)\right)\right). \label{eq_enc_ffn}
\end{align}
$\mathbf{H}^l \in \mathbb{R}^{n\times d}$ denotes the sequence representation of the $l$-th encoder layer. Input to the first layer $\mathbf{H}^0$ is the element-wise addition of the source word embedding $\mathbf{X}$ and the corresponding positional encoding. $\textsc{Ffn}(\cdot)$ is a two-layer feed-forward network with a large intermediate representation and $\text{ReLU}$ activation function. Each encoder sublayer is wrapped with a residual connection (Eq.~\ref{eq_res_con}), followed by layer normalization (Eq.~\ref{eq_ln}):
\begin{align}
    \textsc{Rc}(\mathbf{z}, \mathbf{z}^\prime) & = \mathbf{z} + \mathbf{z}^\prime, \label{eq_res_con} \\
    \textsc{Ln}(\mathbf{z}) & = \frac{\mathbf{z} - \mu}{\sigma} \odot \mathbf{g} + \mathbf{b}, \label{eq_ln}
\end{align}
where $\mathbf{z}$ and $\mathbf{z}^\prime$ are input vectors, and $\odot$ indicates element-wise multiplication. $\mu$ and $\sigma$ denote the mean and standard deviation statistics of vector $\mathbf{z}$. The normalized $\mathbf{z}$ is then re-scaled and re-centered by trainable parameters $\mathbf{g}$ and $\mathbf{b}$ individually. 

The decoder also consists of $L$ identical layers, each of them extends the encoder sublayers with an encoder-decoder attention sublayer (Eq.~\ref{eq_dec_enc_dec_att}) to capture translation alignment from target words to relevant source words:
\begin{align}
\mathbf{\tilde{S}}^l & = \textsc{Ln}\left(\textsc{Rc}\left(\mathbf{S}^{l-1}, \textsc{Att}(\mathbf{S}^{l-1}, \mathbf{S}^{l-1})\right)\right) \label{eq_dec_self_att} \\
\mathbf{\bar{S}}^l & = \textsc{Ln}\left(\textsc{Rc}\left(\mathbf{\tilde{S}}^{l}, \textsc{Att}(\mathbf{\tilde{S}}^{l}, \mathbf{H}^{L})\right)\right) \label{eq_dec_enc_dec_att} \\
\mathbf{S}^l &= \textsc{Ln}\left(\textsc{Rc}\left(\mathbf{\bar{S}}^l, \textsc{Ffn}(\mathbf{\bar{S}}^l)\right)\right) \label{eq_dec_ffn}.
\end{align}
$\mathbf{S}^l \in \mathbb{R}^{m\times d}$ is the sequence representation of the $l$-th decoder layer. Input $\mathbf{S}^0$ is defined similar to $\mathbf{H}^0$. To ensure auto-regressive decoding, the attention weights in Eq.~\ref{eq_dec_self_att} are masked to prevent attention to future target tokens.

The Transformer's parameters are typically initialized by sampling from a uniform distribution:
\begin{equation}
    \mathbf{W} \in \mathbb{R}^{d_{i} \times d_{o}} \sim \mathcal{U}\left(-\gamma, \gamma\right), \gamma = \sqrt{\frac{6}{d_{i}+d_{o}}}, \label{eq_init}
\end{equation}
where $d_i$ and $d_o$ indicate input and output dimension separately. This initialization has the advantage of maintaining activation variances and back-propagated gradients variance and can help train deep neural networks~\cite{article2010golort}.

\section{Vanishing Gradient Analysis}\label{sec_vanish}

One natural way to deepen Transformer is simply enlarging the layer number $L$. Unfortunately, Figure \ref{fig_encdec_gradnorm} shows that this would give rise to gradient vanishing on both the encoder and the decoder at the lower layers, and that the case on the decoder side is worse.
We identified a structural problem in the Transformer architecture that gives rise to this issue, namely the interaction of RC and LN, which we will here discuss in more detail.

Given an input vector $\mathbf{z} \in \mathbb{R}^d$, let us consider the general structure of RC followed by LN:
\begin{align}
    \mathbf{r} & = \textsc{Rc}\left(\mathbf{z}, f(\mathbf{z})\right) \label{eq_g_rc}, \\
    \mathbf{o} & = \textsc{Ln}\left(\mathbf{r}\right) \label{eq_g_ln},
\end{align}
where $\mathbf{r}, \mathbf{o} \in \mathbb{R}^d$ are intermediate outputs. $f(\cdot)$ represents any neural network, such as recurrent, convolutional or attention network, etc. Suppose during back-propagation, the error signal at the output of LN is $\mathbf{\delta}_o$. 
Contributions of RC and LN to the error signal are as follows:
\begin{align}
    \mathbf{\delta}_r & = \frac{\partial \mathbf{o}}{\partial \mathbf{r}} \mathbf{\delta}_o = \text{diag}(\frac{\mathbf{g}}{\sigma_r}) ( \mathbf{I} - \frac{1-\mathbf{\bar{r}}\mathbf{\bar{r}}^T}{d})  \mathbf{\delta}_o \label{eq_partial_g_ln} \\
    \mathbf{\delta}_z & = \frac{\partial \mathbf{r}}{\partial \mathbf{z}} \mathbf{\delta}_r = (1 + \frac{\partial f}{\partial \mathbf{z}}) \mathbf{\delta}_r \label{eq_partial_g_rc},
\end{align}
where $\mathbf{\bar{r}}$ denotes the normalized input.
$\mathbf{I}$ is the identity matrix and $\text{diag}(\cdot)$ establishes a diagonal matrix from its input. The resulting $\mathbf{\delta}_r$ and $\mathbf{\delta}_z$ are error signals arrived at output $\mathbf{r}$ and $\mathbf{z}$ respectively. 

We define the change of error signal as follows:
\begin{equation}
    \beta = \beta_{\textsc{Ln}} \cdot \beta_{\textsc{Rc}} = \frac{\|\delta_z\|_2}{\|\delta_r\|_2} \cdot \frac{\|\delta_r\|_2}{\|\delta_o\|_2}, \label{eq_decay_es}
\end{equation}
where $\beta$ (or model ratio), $\beta_{\textsc{Ln}}$ (or LN ratio) and $\beta_{\textsc{Rc}}$ (or RC ratio) measure the gradient norm ratio\footnote{Model gradients depend on both error signal and layer activation. Reduced/enhanced error signal does not necessarily result in gradient vanishing/explosion, but strongly contributes to it.} of the whole residual block, the layer normalization and the residual connection respectively. Informally, a neural model should preserve the gradient norm between layers ($\beta \approx 1$) so as to allow training of very deep models \cite[see][]{DBLP:journals/corr/abs-1805-07477}.

\begin{table}[t]
\centering
\small
\begin{tabular}{l|c|c|ccc}
Method & Module &  & Self & Cross & FFN \\
\hline
\multirow{8}{*}{Base} & \multirow{4}{*}{Enc}
  & $\beta_{\textsc{Ln}}$ & 0.86 & - & 0.84 \\
& & $\beta_{\textsc{Rc}}$ & 1.22 & - & 1.10 \\
& & $\beta$ & 1.05 & - & 0.93 \\
& & $\text{Var}(\mathbf{r})$ & 1.38 & - & 1.40 \\
\cline{2-6}
& \multirow{4}{*}{Dec}
  & $\beta_{\textsc{Ln}}$ & 0.82 & 0.74 & 0.84 \\
& & $\beta_{\textsc{Rc}}$ & 1.21 & 1.00 & 1.11 \\
& & $\beta$ & 0.98 & 0.74 & 0.93 \\
& & $\text{Var}(\mathbf{r})$ & 1.48 & 1.84 & 1.39 \\
\hline
\multirow{8}{*}{Ours} & \multirow{4}{*}{Enc}
  & $\beta_{\textsc{Ln}}$ & 0.96 & - & 0.95 \\
& & $\beta_{\textsc{Rc}}$ & 1.04 & - & 1.02 \\
& & $\beta$ & 1.02 & - & 0.98 \\
& & $\text{Var}(\mathbf{r})$ & 1.10 & - & 1.10 \\
\cline{2-6}
& \multirow{4}{*}{Dec}
  & $\beta_{\textsc{Ln}}$ & 0.95 & 0.94 & 0.94 \\
& & $\beta_{\textsc{Rc}}$ & 1.05 & 1.00 & 1.02 \\
& & $\beta$ & 1.10 & 0.95 & 0.98 \\
& & $\text{Var}(\mathbf{r})$ & 1.13 & 1.15 & 1.11 \\

\end{tabular}
\caption{\label{tb_factor} Empirical measure of output variance $\text{Var}(\mathbf{r})$ of RC and error signal change ratio $\beta_{\textsc{Ln}}$, $\beta_{\textsc{Rc}}$ and $\beta$ (Eq.~\ref{eq_decay_es}) averaged over 12 layers. These values are estimated with $\sim$3k target tokens at the beginning of training using 12-layer Transformer. ``\textit{Base}'': the baseline Transformer. ``\textit{Ours}'': the Transformer with DS-Init. \textit{Enc} and \textit{Dec} stand for encoder and decoder respectively. \textit{Self}, \textit{Cross} and \textit{FFN} indicate the self-attention, encoder-decoder attention and the feed-forward sublayer respectively.} 
\end{table}
We resort to empirical evidence to analyze these ratios. Results in Table \ref{tb_factor} show that LN weakens error signal ($\beta_{\textsc{Ln}} < 1$) but RC strengthens it ($\beta_{\textsc{Rc}} > 1$). One explanation about LN's decay effect is the large output variance of RC ($\text{Var}(\mathbf{r}) > 1$) which negatively affects $\mathbf{\delta}_r$ as shown in Eq.~\ref{eq_partial_g_ln}. By contrast, the short-cut in RC ensures that the error signal at higher layer $\mathbf{\delta}_r$ can always be safely carried on to lower layer no matter how complex $\frac{\partial f}{\partial \mathbf{z}}$ would be as in Eq.~\ref{eq_partial_g_rc}, increasing the ratio.

\begin{figure*}[t]
  \centering
  \mbox{
    \subfigure[\label{fig_dec_self_att} Self-Attention]{\includegraphics[scale=0.35]{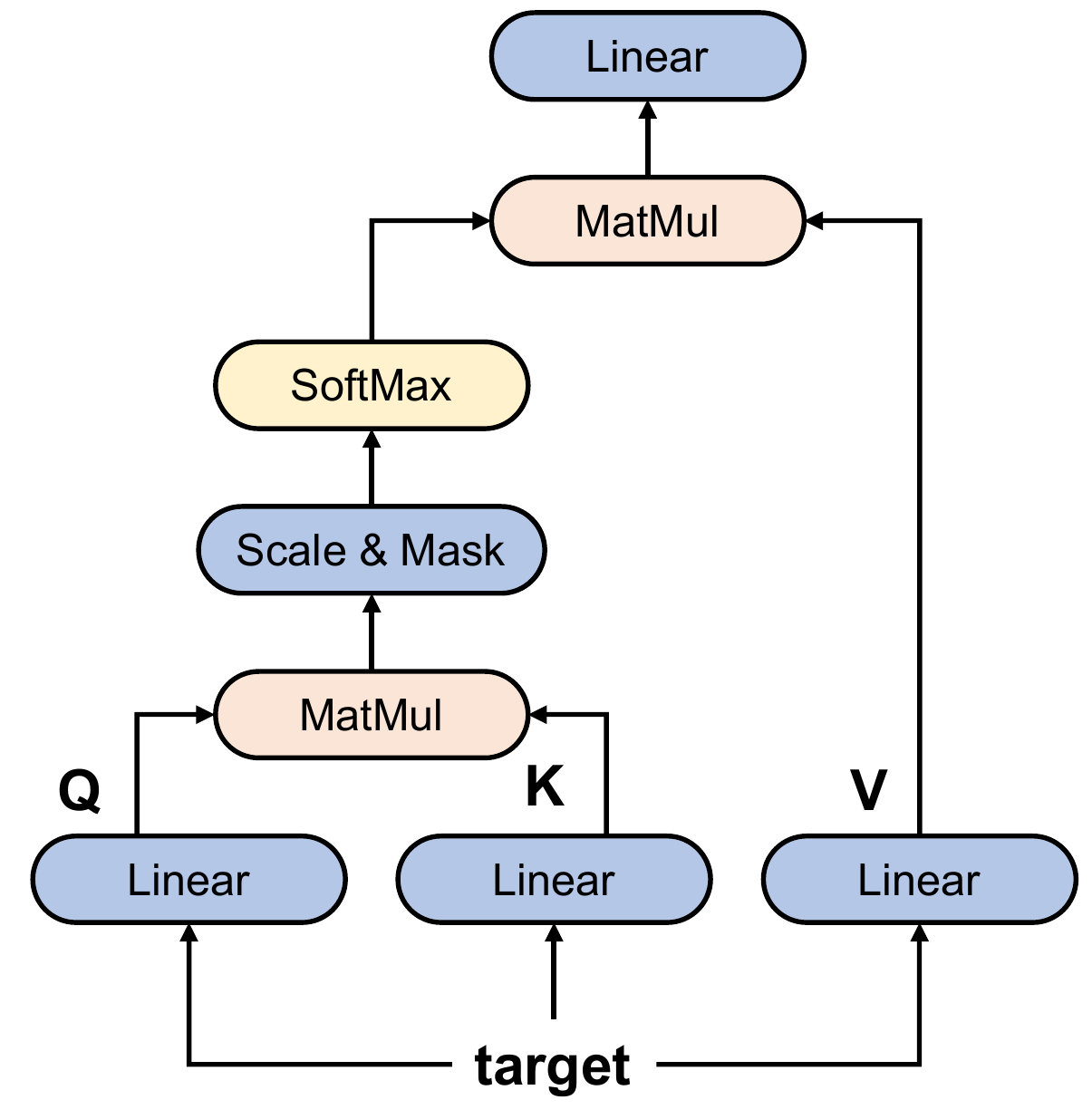}}\quad
    \subfigure[\label{fig_dec_aan} AAN]{\includegraphics[scale=0.35]{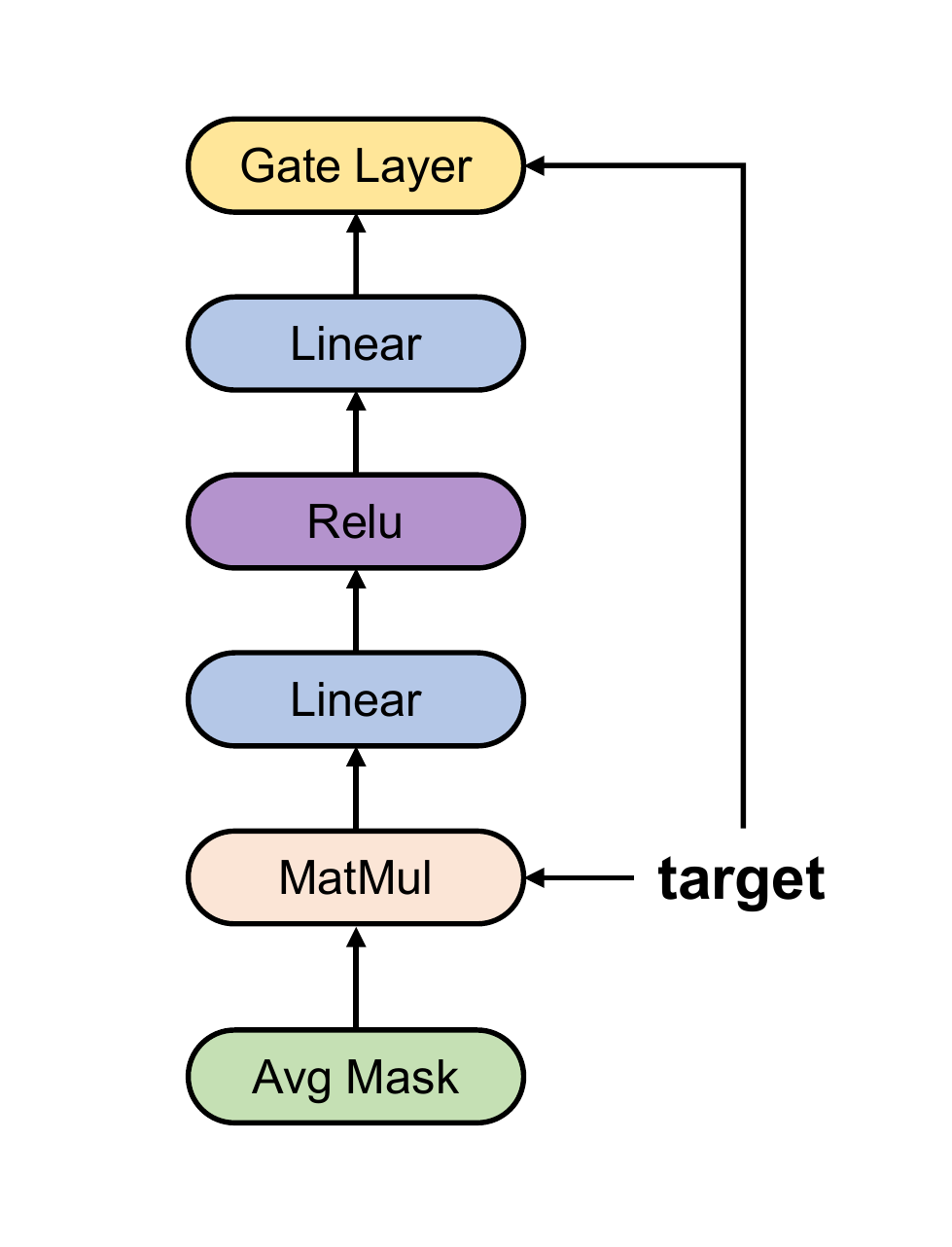}}\quad
    \subfigure[\label{fig_dec_merged_att} Merged attention with simplified AAN]{\includegraphics[scale=0.35]{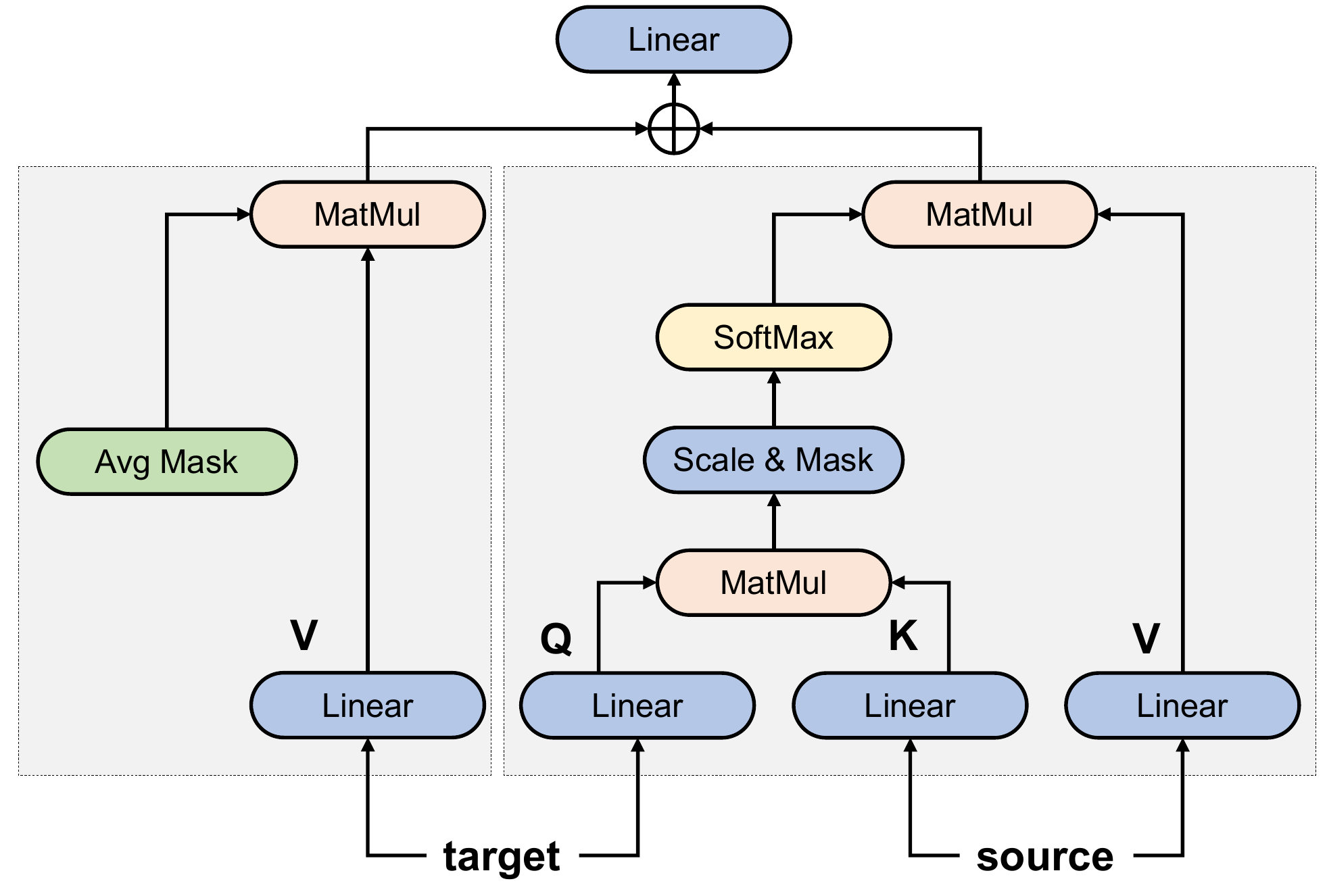}}
  }
  \caption{\label{fig_merged_att} An overview of self-attention, AAN and the proposed merged attention with simplified AAN.}
\end{figure*}
\section{Depth-Scaled Initialization}
Results on the model ratio show that self-attention sublayer has a (near) increasing effect ($\beta > 1$) that intensifies error signal, while feed-forward sublayer manifests a decreasing effect ($\beta < 1$). In particular, though the encoder-decoder attention sublayer and the self-attention sublayer share the same attention formulation, the model ratio of the former is smaller. As shown in Eq.\  \ref{eq_dec_enc_dec_att} and \ref{eq_att}, part of the reason is that encoder-decoder attention can only back-propagate gradients to lower layers through the query representation $\mathbf{Q}$, bypassing gradients at the key $\mathbf{K}$ and the value $\mathbf{V}$ to the encoder side. This negative effect explains why the decoder suffers from more severe gradient vanishing than the encoder in Figure \ref{fig_encdec_gradnorm}.

The gradient norm is preserved better through the self-attention layer than the encoder-decoder attention, which offers insights on the successful training of the deep Transformer in BERT~\cite{devlin2018bert} and GPT~\cite{radford2018improving}, where encoder-decoder attention is not involved. However, results in Table \ref{tb_factor} also suggests that the self-attention sublayer in the encoder is not strong enough to counteract the gradient loss in the feed-forward sublayer. That is why BERT and GPT adopt a much smaller standard deviation (0.02) for initialization, in a similar spirit to our solution. 

We attribute the gradient vanishing issue to the large output variance of RC (Eq.~\ref{eq_partial_g_ln}). Considering that activation variance is positively correlated with parameter variance~\cite{article2010golort}, we propose DS-Init and change the original initialization method in Eq.~\ref{eq_init} as follows:
\begin{equation}
    \mathbf{W} \in \mathbb{R}^{d_{i} \times d_{o}} \sim \mathcal{U}\left(-\gamma \frac{\alpha}{\sqrt{l}}, \gamma \frac{\alpha}{\sqrt{l}}\right), \label{eq_dsinit}
\end{equation}
where $\alpha$ is a hyperparameter in the range of $[0, 1]$ and $l$ denotes layer depth. Hyperparameter $\alpha$ improves the flexibility of our method. Compared with existing approaches~\cite{vaswani2018tensor2tensor,bapna-etal-2018-training}, our solution
does not require modifications in the model architecture and hence is easy to implement. 

According to the property of uniform distribution, the variance of model parameters decreases from $\frac{\gamma^2}{3}$ to $\frac{\gamma^2\alpha^2}{3l}$ after applying DS-Init. By doing so, a higher layer would have smaller output variance of RC so that more gradients can flow back. Results in Table \ref{tb_factor} suggest that DS-Init narrows both the variance and different ratios to be $\sim$1, ensuring the stability of gradient back-propagation. Evidence in Figure \ref{fig_encdec_gradnorm} also shows that DS-Init helps keep the gradient norm and slightly increases it on the encoder side. This is because DS-Init endows lower layers with parameters of larger variance and activations of larger norm. When error signals at different layers are of similar scale, the gradient norm at lower layers would be larger. Nevertheless, this increase does not hurt model training based on our empirical observation.

DS-Init is partially inspired by the \textit{Fixup} initialization~\cite{zhang2018residual}. Both of them try to reduce the output variance of RC. The difference is that Fixup focuses on overcoming gradient explosion cased by consecutive RCs and seeks to enable training without LN but at the cost of carefully handling parameter initialization of each matrix transformation, including manipulating initialization of different bias and scale terms. Instead, DS-Init aims at solving gradient vanishing in deep Transformer caused by the structure of RC followed by LN. We still employ LN to standardize layer activation and improve model convergence. The inclusion of LN ensures the stability and simplicity of DS-Init.

\section{Merged Attention Model}

With large model depth, deep Transformer unavoidably introduces high computational overhead. This brings about significantly longer training and decoding time. To remedy this issue, we propose a merged attention model for decoder that integrates a simplified average-based self-attention sublayer into the encoder-decoder attention sublayer. Figure \ref{fig_merged_att} highlights the difference.

The AAN model (Figure \ref{fig_dec_aan}), as an alternative to the self-attention model (Figure \ref{fig_dec_self_att}), accelerates Transformer decoding by allowing decoding in linear time, avoiding the $\mathcal{O}(n^2)$ complexity of the self-attention mechanism~\cite{zhang-etal-2018-accelerating}. Unfortunately, the gating sublayer and the feed-forward sublayer inside AAN reduce the empirical performance improvement. We propose a simplified AAN by removing all matrix computation except for two linear projections:
\begin{equation}
    \textsc{SAan}(\mathbf{S}^{l-1}) = \left[\mathbf{M}_a (\mathbf{S}^{l-1} \mathbf{W}_v)\right] \mathbf{W}_o,
\end{equation}
where $\mathbf{M}_a$ denotes the average mask matrix for parallel computation~\cite{zhang-etal-2018-accelerating}. This new model is then combined with the encoder-decoder attention as shown in Figure \ref{fig_dec_merged_att}:
\begin{align}
    & \textsc{MAtt}(\mathbf{S}^{l-1}) = \textsc{SAan}(\mathbf{S}^{l-1}) + \textsc{Att}(\mathbf{S}^{l-1}, \mathbf{H}^L) \nonumber \\
    & \mathbf{\bar{S}}^l  = \textsc{Ln}\left(\textsc{Rc}\left(\mathbf{{S}}^{l-1}, \textsc{MAtt}(\mathbf{{S}}^{l-1})\right)\right). \label{eq_merged_att}
\end{align}
The mapping $\mathbf{W}_o$ is shared for $\textsc{SAan}$ and $\textsc{Att}$. After combination, MAtt allows for the parallelization of AAN and encoder-decoder attention.

\section{Experiments}

\begin{table}[t]
\centering
\small
\begin{tabular}{l|rrrc}
Dataset & \#Src & \#Tgt & \#Sent & \#BPE  \\
\hline
WMT14 En-De & 116M  & 110M & 4.5M & 32K \\
WMT14 En-Fr & 1045M & 1189M & 36M & 32K \\
WMT18 En-Fi & 73M & 54M & 3.3M & 32K \\
WMT18 Zh-En & 510M & 576M  & 25M & 32K \\
IWSLT14 De-En & 3.0M & 3.2M & 159K & 30K \\
\end{tabular}
\caption{\label{tb_dataset} Statistics for different training datasets. \textit{\#Src} and \textit{\#Tgt} denote the number of source and target tokens respectively. \textit{\#Sent}: the number of bilingual sentences. \textit{\#BPE}: the number of merge operations in BPE. \textit{M}: million, \textit{K}: thousand.} 
\end{table}
\begin{table*}[t]
\centering
\small
\begin{tabular}{c|l|c|c|Hc|c}
ID & Model & \#Param & Test14 & Test14-18 & $\bigtriangleup$Dec & $\bigtriangleup$Train \\
\hline
1 & Base 6 layers $dp_a=dp_r=0.1,bs=25K$ & 72.3M  & 27.59 (26.9) & 32.47 (32.20) & {62.26}/1.00$\times$ & {0.105}/1.00$\times$  \\
2 & 1 + T2T & 72.3M & 27.20 (26.5) & 31.88 (31.62) & {68.04}/0.92$\times$ & {0.105}/1.00$\times$ \\
\hline
3 & 1 + DS-Init & 72.3M & 27.50 (26.8) & 32.26 (31.96) & {$\star$}/1.00$\times$ & {$\star$}/1.00$\times$ \\
4 & 1 + MAtt & 66.0M & 27.49 (26.8) & 31.96 (31.68) & {40.51}/1.54$\times$ & {0.094}/1.12$\times$ \\
5 & 1 + MAtt + DS-Init & 66.0M & 27.35 (26.8) & 32.01 (31.76) & {40.84}/1.52$\times$ & {0.094}/1.12$\times$ \\
\hline
6 & 1 + MAtt with self-attention & 72.3M & 27.41 (26.7) & 32.07 (31.78) & {60.25}/1.03$\times$ & {0.105}/1.00$\times$ \\
7 & 1 + MAtt with original AAN & 72.2M & 27.36 (26.7) & 32.11 (31.82) & {46.13}/1.35$\times$ & {0.098}/1.07$\times$ \\
8 & 1 + $bs=50K$ & 72.3M & 27.84 (27.2) & 32.38 (32.10) & {$\star$}/1.00$\times$ & {$\star$}/1.00$\times$ \\
\hline
9 & 1 + $>12$ layers + $bs=25K\sim 50K$ & - & - & - & - & - \\
10 & 4 + $>12$ layers + $bs=25K\sim 50K$ & - & - & - & - \\
\hline
11 & 3 + 12 layers + $bs=40K,dp_r=0.3,dp_a=0.2$ & 116.4M & 28.27 (27.6) & 32.78 (32.48) & {102.9}/1.00$\times^\dagger$ & {0.188}/1.00$\times^\dagger$ \\
12 & 11 + T2T & 116.5M & 28.03 (27.4) & \textbf{32.99} (32.64) & {107.7}/0.96$\times^\dagger$ & {0.191}/0.98$\times^\dagger$ \\
13 & 11 + MAtt & 103.8M & 28.55 (27.9) & 32.87 (32.58) & {67.12}/1.53$\times^\dagger$ & {0.164} /1.15$\times^\dagger$ \\
\hline
14 & 3 + 20 layers + $bs=44K,dp_r=0.3,dp_a=0.2$ & 175.3M & 28.42 (27.7) & 32.75 (32.46) & {157.8}/1.00$\times^\ddagger$ &  {0.283}/1.00$\times^\ddagger$ \\
15 & 14 + T2T & 175.3M & 28.27 (27.6) & 32.80 (32.50) & {161.2}/0.98$\times^\ddagger$ & {0.289}/0.98$\times^\ddagger$ \\
16 & 14 + MAtt & 154.3M & \textbf{28.67} (\textbf{28.0}) & 32.92 (\textbf{32.66}) & {108.6}/1.45$\times^\ddagger$ & {0.251}/1.13$\times^\ddagger$ \\
\end{tabular}
\caption{\label{tb_wmt14_en_de} Tokenized case-sensitive BLEU (in parentheses: sacreBLEU) on WMT14 En-De translation task.  \textit{\#Param}: number of model parameters. \textit{$\bigtriangleup$Dec}: decoding time (seconds)/speedup on newstest2014 dataset with a batch size of 32. \textit{$\bigtriangleup$Train}: training time (seconds)/speedup per training step evaluated on 0.5K steps with a batch size of 1K target tokens. Time is averaged over 3 runs using Tensorflow on a single TITAN X (Pascal). ``-'': optimization failed and no result. {``$\star$'': the same as model \textcircled{1}.} $^\dagger$ and $^\ddagger$: comparison against \textcircled{11} and \textcircled{14} respectively rather than \textcircled{1}. \textit{Base}: the baseline Transformer with base setting. Bold indicates best BLEU score. $dp_a$ and $dp_r$: dropout rate on attention weights and residual connection. $bs$: batch size in tokens.
} 
\end{table*}

\subsection{Datasets and Evaluation}\label{sec_dataset}

We take WMT14 English-German translation (En-De) \cite{bojar-EtAl:2014:W14-33} as our benchmark for model analysis, and examine the generalization of our approach on four other tasks: WMT14 English-French (En-Fr), IWSLT14 German-English (De-En) \cite{iwslt2014}, WMT18 English-Finnish (En-Fi) and WMT18 Chinese-English (Zh-En) \cite{bojar-etal-2018-findings}. Byte pair encoding algorithm (BPE)~\cite{sennrich-etal-2016-neural} is used in preprocessing to handle low frequency words. Statistics of different datasets are listed in Table \ref{tb_dataset}. 

Except for IWSLT14 De-En task, we collect subword units independently on the source and target side of training data. We directly use the preprocessed training data from the WMT18 website\footnote{http://www.statmt.org/wmt18/translation-task.html} for En-Fi and Zh-En tasks, and use newstest2017 as our development set, newstest2018 as our test set. Our training data for WMT14 En-De and WMT14 En-Fr is identical to previous setups~\cite{NIPS2017_7181,wu2018pay}. We use newstest2013 as development set for WMT14 En-De and newstest2012+2013 for WMT14 En-Fr. Apart from newstest2014 test set\footnote{We use the filtered test set consisting of 2737 sentence pairs. The difference of translation quality on filtered and full test sets is marginal.}, we also evaluate our model on all WMT14-18 test sets for WMT14 En-De translation. The settings for IWSLT14 De-En are as in \citet{DBLP:journals/corr/RanzatoCAZ15}, with 7584 sentence pairs for development, and the concatenated dev sets for IWSLT 2014 as test set (tst2010, tst2011, tst2012, dev2010, dev2012).

We report tokenized case-sensitive BLEU~\cite{Papineni:2002:BMA:1073083.1073135} for WMT14 En-De and WMT14 En-Fr, and provide detokenized case-sensitive BLEU for WMT14 En-De, WMT18 En-Fi and Zh-En with \textit{sacreBLEU}~\cite{W18-6319}\footnote{Signature BLEU+c.mixed+\#.1+s.exp+tok.13a+v.1.2.20}. We also report chrF score for En-Fi translation which was found correlated better with human evaluation~\cite{bojar-etal-2018-findings}. Following previous work~\cite{wu2018pay}, we evaluate IWSLT14 De-En with tokenized case-insensitive BLEU.

\subsection{Model Settings}

We experiment with both \textit{base} (layer size 512/2048, 8 heads) and \textit{big} (layer size 1024/4096, 16 heads) settings as in \citet{NIPS2017_7181}. Except for the vanilla Transformer, we also compare with the structure that is currently default in tensor2tensor (T2T), which puts layer normalization before residual blocks~\cite{vaswani2018tensor2tensor}. We use an in-house toolkit for all experiments. 

Dropout is applied to the residual connection ($dp_r$) and attention weights ($dp_a$). We share the target embedding matrix with the softmax projection matrix but not with the source embedding matrix. We train all models using Adam optimizer (0.9/0.98 for base, 0.9/0.998 for big) with adaptive learning rate schedule (warm-up step 4K for base, 16K for big) as in~\cite{NIPS2017_7181} and label smoothing of 0.1. We set $\alpha$ in DS-Init to 1.0. Sentence pairs containing around 25K$\sim$50K ($bs$) target tokens are grouped into one batch. We use relatively larger batch size and dropout rate for deeper and bigger models for better convergence. We perform evaluation by averaging last 5 checkpoints. Besides, we apply mixed-precision training to all big models. Unless otherwise stated, we train base and big model with 300K maximum steps, and decode sentences using beam search with a beam size of 4 and length penalty of 0.6. Decoding is implemented with cache to save redundant computations. Other settings for specific translation tasks are explained in the individual subsections.

\subsection{WMT14 En-De Translation Task}

Table \ref{tb_wmt14_en_de} summarizes translation results under different settings. Applying DS-Init and/or MAtt to Transformer with 6 layers slightly decreases translation quality by $\sim$0.2 BLEU (27.59$\rightarrow$27.35). However, they allow scaling up to deeper architectures, achieving a BLEU score of 28.55 (12 layers) and 28.67 (20 layers), outperforming all baselines. These improvements can not be obtained via enlarging the training batch size (\textcircled{8}), confirming the strength of deep models.

We also compare our simplified AAN in MAtt (\textcircled{4}) with two variants: a self-attention network (\textcircled{6}), and the original AAN (\textcircled{7}). Results show minor differences in translation quality, but improvements in training and decoding speed, and a reduction in the number of model parameters.
Compared to the baseline, MAtt improves decoding speed by 50\%, and training speed by 10\%, while having 9\% fewer parameters.

Result \textcircled{9} indicates that the gradient vanishing issue prevents training of deep vanilla Transformers, which cannot be solved by only simplifying the decoder via MAtt (\textcircled{10}). By contrast, both T2T and DS-Init can help. Our DS-Init improves norm preservation through specific parameter initialization, while T2T reschedules the LN position. Results in Table \ref{tb_wmt14_en_de} show that T2T underperforms DS-Init by 0.2 BLEU on average, and slightly increases training and decoding time (by 2\%) compared to the original Transformer due to additional LN layers. This suggests that our solution is more effective and efficient.

\begin{table}[t]
\centering
\small
\begin{tabular}{l|cc|cc}
\multirow{2}{*}{ID}& \multicolumn{2}{c|}{BLEU} & \multicolumn{2}{c}{PPL} \\
\cline{2-5}
& Train & Dev & Train & Dev \\
\hline
1 & 28.64 & 26.16 & 5.23 & 4.76 \\
\hline
11 & 29.63 & 26.44 & \textbf{4.48} & \textbf{4.38} \\
12 & \textbf{29.75} & 26.16 & 4.60 & 4.49 \\
13 & 29.43 & \textbf{26.51} & 5.09 & 4.71 \\
\hline
14 & 30.71 & 26.52 & \textbf{3.96} & \textbf{4.32} \\
15 & \textbf{30.89} & 26.53 & 4.09 & 4.41 \\
16 & 30.25 & \textbf{26.56} & 4.62 & 4.58 \\
\end{tabular}
\caption{\label{tb_why_matt_good} Tokenized case-sensitive BLEU (BLEU) and perplexity (PPL) on training (Train) and development (newstest2013, Dev) set. We randomly select 3K sentence pairs as our training data for evaluation. Lower PPL is better.} 
\end{table}

Surprisingly, training deep Transformers with both DS-Init and MAtt improves not only running efficiency but also translation quality (by 0.2 BLEU), compared with DS-Init alone. To get an improved understanding, we analyze model performance on both training and development set. Results in Table \ref{tb_why_matt_good} show that models with DS-Init yield the best perplexity on both training and development set, and those with T2T achieve the best BLEU on the training set. However, DS-Init$+$MAtt performs best in terms of BLEU on the development set. This indicates that the success of DS-Init$+$MAtt comes from its better generalization rather than better fitting training data.

\begin{table*}[t]
\centering
\small
\begin{tabular}{l|l|c|c|c|c}
Task & Model & \#Param & BLEU & $\bigtriangleup$Dec & $\bigtriangleup$Train \\
\hline
\multirow{2}{*}{WMT14 En-Fr} 
    & Base + 6 layers & 76M & 39.09 & 167.56/1.00$\times$ & 0.171/1.00$\times$ \\
    & Ours + Base + 12 layers & 108M & \textbf{40.58} & 173.62/0.97$\times$ & 0.265/0.65$\times$ \\
\hline
\multirow{2}{*}{IWSLT14 De-En}
    & Base + 6 layers & 61M & 34.41 & 315.59/1.00$\times$ & 0.153/1.00$\times$ \\
    & Ours + Base + 12 layers & 92M & \textbf{35.63} & 329.95/0.96$\times$ & 0.247/0.62$\times$ \\
\hline
\multirow{2}{*}{WMT18 En-Fi}
    & Base + 6 layers & 65M & 15.5 (50.82) & 156.32/1.00$\times$ & 0.165/1.00$\times$ \\
    & Ours + Base + 12 layers & 96M & \textbf{15.8} (\textbf{51.47}) & 161.74/0.97$\times$ & 0.259/0.64$\times$ \\
\hline
\multirow{2}{*}{WMT18 Zh-En}
    & Base + 6 layers & 77M & 21.1 & 217.40/1.00$\times$ & 0.173/1.00$\times$ \\
    & Ours + Base + 12 layers & 108M & \textbf{22.3} & 228.57/0.95$\times$ & 0.267/0.65$\times$
\end{tabular}
\caption{\label{tb_other_mt} Translation results on different tasks. Settings for BLEU score is given in Section \ref{sec_dataset}. Numbers in bracket denote chrF score. Our model outperforms the vanilla base Transformer on all tasks.  ``\textit{Ours}'': DS-Init$+$MAtt.}
\end{table*}

\begin{figure}[t]
  \centering
    \includegraphics[scale=0.45]{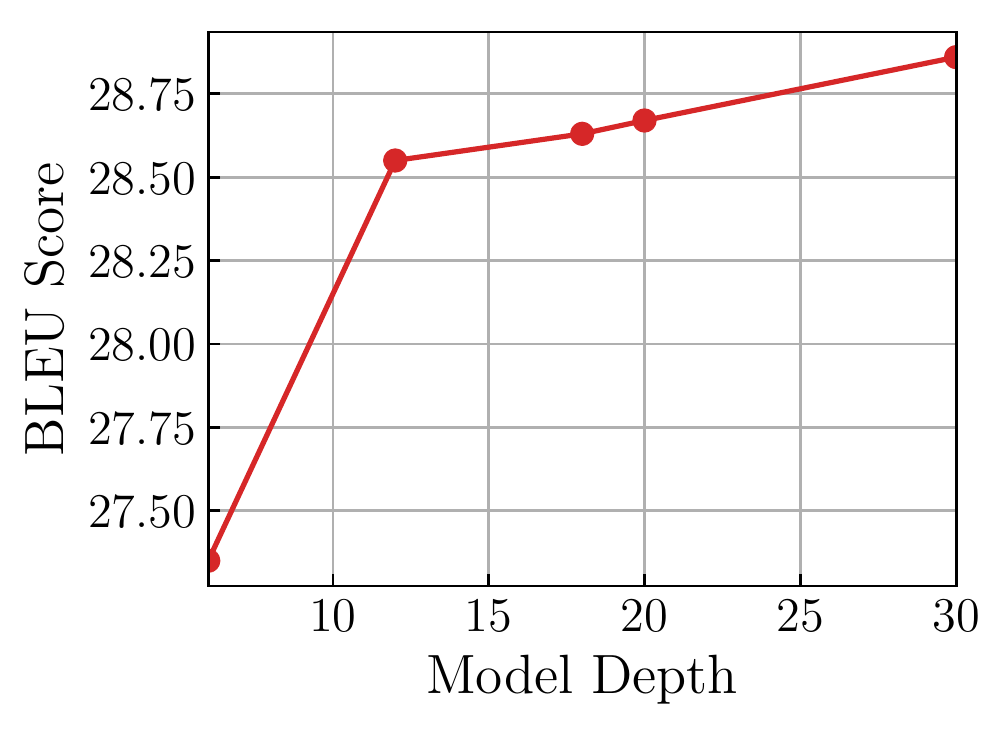}
  \caption{\label{fig_depth_bleu_acc} Test BLEU score on newstest2014 with respect to model depth for Transformer$+$DS-Init$+$MAtt.}
\end{figure}

\begin{table}[t]
\centering
\small
\begin{tabular}{l|c|c|c}
Model & \#Param & Test14 & Test14-18 \\
\hline
\citet{NIPS2017_7181} & 213M & 28.4 & - \\
\citet{P18-1008} & 379M & 28.9 & - \\
\citet{ott2018scaling} & 210M & 29.3 & - \\
\citet{bapna-etal-2018-training} & 137M & 28.04 & - \\
\citet{wu2018pay} & 213M & \makecell{\textbf{29.76} \\ (29.0)}$^\ast$ & \makecell{33.13 \\ (32.86)}$^\ast$ \\
\hline
Big + 6 layers & 233M & \makecell{29.07 \\ (28.3)} & \makecell{33.16 \\ (32.88)} \\
Ours + Big + 12 layers & 359M & \makecell{29.47 \\ (28.7)} & \makecell{33.21 \\ (32.90)} \\ 
Ours + Big + 20 layers & 560M & \makecell{29.62 \\ (29.0)} & \makecell{\textbf{33.26} \\ (\textbf{32.96})} \\
\end{tabular}
\caption{\label{tb_wmt14_sota_cmp} Tokenized case-sensitive BLEU (sacreBLEU) on WMT14 En-De translation task. ``\textit{Test14-18}'': BLEU score averaged over newstest2014$\sim$newstest2018. $^\ast$: results obtained by running code and model released by \citet{wu2018pay}. ``\textit{Ours}'': DS-Init$+$MAtt.} 
\end{table}

We also attempt to apply DS-Init on the encoder alone or the decoder alone for 12-layer models. Unfortunately, both variants lead to unstable optimization where gradients tend to explode at the middle of training. We attempt to solve this issue with gradient clipping of rate 1.0. Results show that this fails for decoder and achieves only 27.89 BLEU for encoder, losing 0.66 BLEU compared with the full variant (28.55). We leave further analysis to future work and recommend using DS-Init on both the encoder and the decoder.

\textit{Effect of Model Depth} We empirically compare a wider range of model depths for Transformer$+$DS-Init$+$MAtt with up to 30 layers. Hyperparameters are the same as for \textcircled{14} except that we use 42K and 48K batch size for 18 and 30 layers respectively. Figure \ref{fig_depth_bleu_acc} shows that deeper Transformers yield better performance. However, improvements are steepest going from 6 to 12 layers, and further improvements are small. 

\subsubsection{Comparison with Existing Work}

Table \ref{tb_wmt14_sota_cmp} lists the results in big setting and compares with current SOTA. Big models are trained with $dp_a=0.1$ and $dp_r=0.3$. The 6-layer baseline and the deeper ones are trained with batch size of 48K and 54K respectively. Deep Transformer with our method outperforms its 6-layer counterpart by over 0.4 points on newstest2014 and around 0.1 point on newstest2014$\sim$newstest2018. Our model outperforms the transparent model~\cite{bapna-etal-2018-training} (+1.58 BLEU), an approach for the deep encoder. Our model performs on par with current SOTA, the dynamic convolution model (DCNN)~\cite{wu2018pay}. In particular, though DCNN achieves encouraging performance on newstest2014, it falls behind the baseline on other test sets. By contrast, our model obtains more consistent performance improvements.

In work concurrent to ours, \newcite{wang-etal-2019-learning} discuss how the placement of layer normalization affects deep Transformers, and compare the original \textit{post-norm} (which we consider our baseline) and a \textit{pre-norm} layout (which we call T2T). Their results also show that pre-norm allows training of deeper Transformers. Our results show that deep post-norm Transformers are also trainable with appropriate initialization, and tend to give slightly better results.

\subsection{Results on Other Translation Tasks}

We use 12 layers for our model in these tasks. We enlarge the dropout rate to $dp_a=0.3, dp_r=0.5$ for IWSLT14 De-En task and train models on WMT14 En-Fr and WMT18 Zh-En with 500K steps. Other models are trained with the same settings as in WMT14 En-De. 

We report translation results on other tasks in Table \ref{tb_other_mt}. Results show that our model beats the baseline on all tasks with gains of over 1 BLEU, except the WMT18 En-Fi where our model yields marginal BLEU improvements (+0.3 BLEU). We argue that this is due to the rich morphology of Finnish, and BLEU's inability to measure improvements below the word level. We also provide the chrF score in which our model gains 0.6 points. In addition, speed measures show that though our model consumes 50+\% more training time, there is only a small difference with respect to decoding time thanks to MAtt. 

\subsection{Analysis of Training Dynamics}

\begin{figure}[t]
\centering
\includegraphics[scale=0.55]{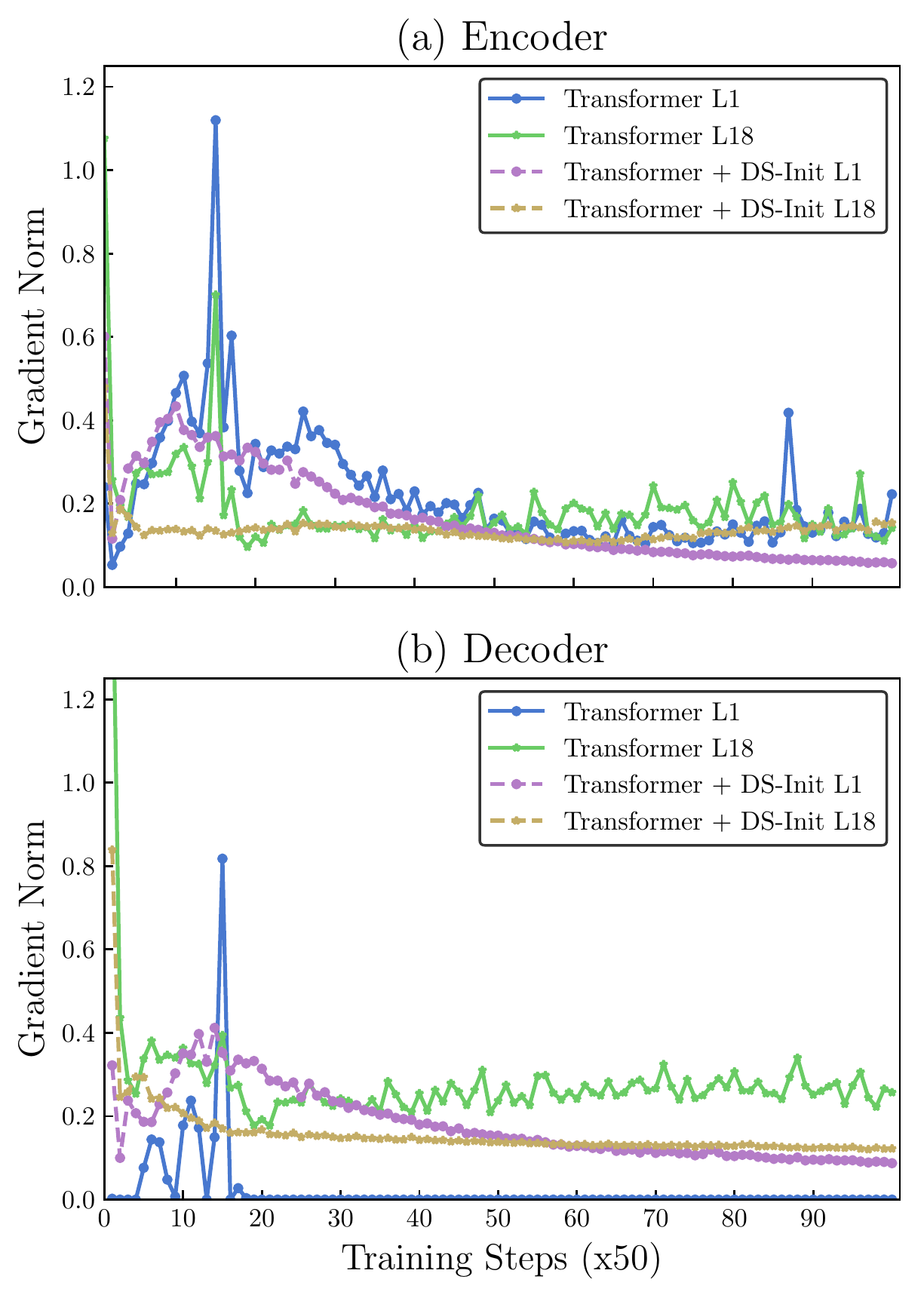}
\caption{\label{fig_train_dynamics} Gradient norm (y-axis) of the first and the last encoder layers (top) and decoder layers (bottom) in 18-layer deep Transformer over the fist 5k training steps. We use around 25k source/target tokens in each training batch. Each point in this plot is averaged over 50 training steps. ``\textit{L1}/\textit{L18}'' denotes the first/last layer. DS-Init helps stabilize the gradient norm during training.}
\end{figure}

Our analysis in Figure \ref{fig_encdec_gradnorm} and Table \ref{tb_factor} is based on gradients estimated exactly after parameter initialization without considering training dynamics. Optimizers with adaptive step rules, such as Adam, could have an adverse effect that enables gradient scale correction through the accumulated first and second moments. However, results in Figure \ref{fig_train_dynamics} show that without DS-Init, the encoder gradients are less stable and the decoder gradients still suffer from the vanishing issue, particularly at the first layer. DS-Init makes the training more stable and robust.\footnote{We observe this both in the raw gradients and after taking the Adam step rules into account.}

\section{Conclusion and Future Work}

This paper discusses training of very deep Transformers. 
We show that the training of deep Transformers suffers from gradient vanishing, which we mitigate with depth-scaled initialization.
To improve training and decoding efficiency, we propose a merged attention sublayer that integrates a simplified average-based self-attention sublayer into the encoder-decoder attention sublayer. Experimental results show that deep models trained with these techniques clearly outperform a vanilla Transformer with 6 layers in terms of BLEU, and outperforms other solutions to train deep Transformers \cite{bapna-etal-2018-training,vaswani2018tensor2tensor}.
Thanks to the more efficient merged attention sublayer, we achieve these quality improvements while matching the decoding speed of the baseline model.

In the future, we would like to extend our model to other sequence-to-sequence tasks, such as summarization and dialogue generation, as well as adapt the idea to other generative architectures~\cite{zhang-etal-2016-variational-neural,zhang2018asynchronous}.
We have trained models with up to 30 layers each for the encoder and decoder, and while training was successful and improved over shallower counterparts, gains are relatively small beyond 12 layers. An open question is whether there are other structural issues that limit the benefits of increasing the depth of the Transformer architecture, or whether the benefit of very deep models is greater for other tasks and dataset.

\section*{Acknowledgments}

We thank the reviewers for their insightful comments.
This project has received funding from the grant H2020-ICT-2018-2-825460 (ELITR) by the European Union.
Biao Zhang also acknowledges the support of the Baidu Scholarship. This work has been performed using resources provided by the Cambridge Tier-2 system operated by the University of Cambridge Research Computing Service (http://www.hpc.cam.ac.uk) funded by EPSRC Tier-2 capital grant EP/P020259/1.

\bibliography{emnlp-ijcnlp-2019}

\begin{thebibliography}{43}
\expandafter\ifx\csname natexlab\endcsname\relax\def\natexlab#1{#1}\fi

\bibitem[{Ba et~al.(2016)Ba, Kiros, and Hinton}]{lei2016layer}
Jimmy~Lei Ba, Jamie~Ryan Kiros, and Geoffrey~E Hinton. 2016.
\newblock Layer normalization.
\newblock \emph{arXiv preprint arXiv:1607.06450}.

\bibitem[{Bahdanau et~al.(2015)Bahdanau, Cho, and
  Bengio}]{bahdanau+al-2014-nmt}
Dzmitry Bahdanau, Kyunghyun Cho, and Yoshua Bengio. 2015.
\newblock {Neural Machine Translation by Jointly Learning to Align and
  Translate}.
\newblock In \emph{{Proceedings of the International Conference on Learning
  Representations (ICLR)}}.

\bibitem[{Bapna et~al.(2018)Bapna, Chen, Firat, Cao, and
  Wu}]{bapna-etal-2018-training}
Ankur Bapna, Mia Chen, Orhan Firat, Yuan Cao, and Yonghui Wu. 2018.
\newblock \href {https://www.aclweb.org/anthology/D18-1338} {Training deeper
  neural machine translation models with transparent attention}.
\newblock In \emph{Proceedings of the 2018 Conference on Empirical Methods in
  Natural Language Processing}, pages 3028--3033, Brussels, Belgium.
  Association for Computational Linguistics.

\bibitem[{Bojar et~al.(2014)Bojar, Buck, Federmann, Haddow, Koehn, Leveling,
  Monz, Pecina, Post, Saint-Amand, Soricut, Specia, and
  Tamchyna}]{bojar-EtAl:2014:W14-33}
Ondrej Bojar, Christian Buck, Christian Federmann, Barry Haddow, Philipp Koehn,
  Johannes Leveling, Christof Monz, Pavel Pecina, Matt Post, Herve Saint-Amand,
  Radu Soricut, Lucia Specia, and Ale\v{s} Tamchyna. 2014.
\newblock \href {http://www.aclweb.org/anthology/W/W14/W14-3302} {Findings of
  the 2014 workshop on statistical machine translation}.
\newblock In \emph{Proceedings of the Ninth Workshop on Statistical Machine
  Translation}, pages 12--58, Baltimore, Maryland, USA. Association for
  Computational Linguistics.

\bibitem[{Bojar et~al.(2018)Bojar, Federmann, Fishel, Graham, Haddow, Koehn,
  and Monz}]{bojar-etal-2018-findings}
Ond{\v{r}}ej Bojar, Christian Federmann, Mark Fishel, Yvette Graham, Barry
  Haddow, Philipp Koehn, and Christof Monz. 2018.
\newblock \href {https://www.aclweb.org/anthology/W18-6401} {Findings of the
  2018 conference on machine translation ({WMT}18)}.
\newblock In \emph{Proceedings of the Third Conference on Machine Translation:
  Shared Task Papers}, pages 272--303, Belgium, Brussels. Association for
  Computational Linguistics.

\bibitem[{Britz et~al.(2017)Britz, Goldie, Luong, and
  Le}]{britz-etal-2017-massive}
Denny Britz, Anna Goldie, Minh-Thang Luong, and Quoc Le. 2017.
\newblock \href {https://doi.org/10.18653/v1/D17-1151} {Massive exploration of
  neural machine translation architectures}.
\newblock In \emph{Proceedings of the 2017 Conference on Empirical Methods in
  Natural Language Processing}, pages 1442--1451, Copenhagen, Denmark.
  Association for Computational Linguistics.

\bibitem[{Cettolo et~al.(2014)Cettolo, Niehues, St{\"u}ker, Bentivogli, and
  Federico}]{iwslt2014}
Mauro Cettolo, Jan Niehues, Sebastian St{\"u}ker, Luisa Bentivogli, and
  Marcello Federico. 2014.
\newblock {Report on the 11th IWSLT Evaluation Campaign, IWSLT 2014}.
\newblock In \emph{{Proceedings of the 11th Workshop on Spoken Language
  Translation}}, pages 2--16, Lake Tahoe, CA, USA.

\bibitem[{Chen et~al.(2018)Chen, Firat, Bapna, Johnson, Macherey, Foster,
  Jones, Schuster, Shazeer, Parmar, Vaswani, Uszkoreit, Kaiser, Chen, Wu, and
  Hughes}]{P18-1008}
Mia~Xu Chen, Orhan Firat, Ankur Bapna, Melvin Johnson, Wolfgang Macherey,
  George Foster, Llion Jones, Mike Schuster, Noam Shazeer, Niki Parmar, Ashish
  Vaswani, Jakob Uszkoreit, Lukasz Kaiser, Zhifeng Chen, Yonghui Wu, and
  Macduff Hughes. 2018.
\newblock \href {http://aclweb.org/anthology/P18-1008} {The best of both
  worlds: Combining recent advances in neural machine translation}.
\newblock In \emph{Proceedings of the 56th Annual Meeting of the Association
  for Computational Linguistics (Volume 1: Long Papers)}, pages 76--86.
  Association for Computational Linguistics.

\bibitem[{Child et~al.(2019)Child, Gray, Radford, and
  Sutskever}]{Child2019GeneratingLS}
Rewon Child, Scott Gray, Alec Radford, and Ilya Sutskever. 2019.
\newblock Generating long sequences with sparse transformers.
\newblock \emph{CoRR}, abs/1904.10509.

\bibitem[{Cho et~al.(2014)Cho, van Merrienboer, Gulcehre, Bahdanau, Bougares,
  Schwenk, and Bengio}]{cho-EtAl:2014:EMNLP2014}
Kyunghyun Cho, Bart van Merrienboer, Caglar Gulcehre, Dzmitry Bahdanau, Fethi
  Bougares, Holger Schwenk, and Yoshua Bengio. 2014.
\newblock {Learning Phrase Representations using RNN Encoder--Decoder for
  Statistical Machine Translation}.
\newblock In \emph{{Proceedings of the 2014 Conference on Empirical Methods in
  Natural Language Processing (EMNLP)}}, pages 1724--1734, Doha, Qatar.

\bibitem[{Devlin et~al.(2019)Devlin, Chang, Lee, and
  Toutanova}]{devlin2018bert}
Jacob Devlin, Ming-Wei Chang, Kenton Lee, and Kristina Toutanova. 2019.
\newblock \href {https://doi.org/10.18653/v1/N19-1423} {{BERT}: Pre-training of
  deep bidirectional transformers for language understanding}.
\newblock In \emph{Proceedings of the 2019 Conference of the North {A}merican
  Chapter of the Association for Computational Linguistics: Human Language
  Technologies, Volume 1 (Long and Short Papers)}, pages 4171--4186,
  Minneapolis, Minnesota. Association for Computational Linguistics.

\bibitem[{Di~Gangi and Federico(2018)}]{di2018deep}
Mattia~Antonino Di~Gangi and Marcello Federico. 2018.
\newblock Deep neural machine translation with weakly-recurrent units.
\newblock In \emph{Proceedings of EAMT}, Alicante, Spain.

\bibitem[{Gehring et~al.(2017)Gehring, Auli, Grangier, Yarats, and
  Dauphin}]{pmlr-v70-gehring17a}
Jonas Gehring, Michael Auli, David Grangier, Denis Yarats, and Yann~N. Dauphin.
  2017.
\newblock \href {http://proceedings.mlr.press/v70/gehring17a.html}
  {Convolutional sequence to sequence learning}.
\newblock In \emph{Proceedings of the 34th International Conference on Machine
  Learning}, volume~70 of \emph{Proceedings of Machine Learning Research},
  pages 1243--1252, International Convention Centre, Sydney, Australia. PMLR.

\bibitem[{Gers and Schmidhuber(2001)}]{Gers:00icannga}
Felix~A. Gers and J{\"u}rgen Schmidhuber. 2001.
\newblock {{Long Short-Term Memory} Learns Context Free and Context Sensitive
  Languages}.
\newblock In \emph{{Proceedings of the {ICANNGA} 2001 Conference}}, volume~1,
  pages 134--137.

\bibitem[{Ghazvininejad et~al.(2019)Ghazvininejad, Levy, Liu, and
  Zettlemoyer}]{Ghazvininejad2019ConstantTimeMT}
Marjan Ghazvininejad, Omer Levy, Yinhan Liu, and Luke~S. Zettlemoyer. 2019.
\newblock Constant-time machine translation with conditional masked language
  models.
\newblock \emph{CoRR}, abs/1904.09324.

\bibitem[{Glorot and Bengio(2010)}]{article2010golort}
Xavier Glorot and Y~Bengio. 2010.
\newblock Understanding the difficulty of training deep feedforward neural
  networks.
\newblock \emph{Journal of Machine Learning Research - Proceedings Track},
  9:249--256.

\bibitem[{Gu et~al.(2018)Gu, Bradbury, Xiong, Li, and
  Socher}]{Gu2018NonAutoregressiveNM}
Jiatao Gu, James Bradbury, Caiming Xiong, Victor~O.K. Li, and Richard Socher.
  2018.
\newblock \href {https://openreview.net/forum?id=B1l8BtlCb} {Non-autoregressive
  neural machine translation}.
\newblock In \emph{International Conference on Learning Representations}.

\bibitem[{He et~al.(2015)He, Zhang, Ren, and Sun}]{DBLP:journals/corr/HeZRS15}
Kaiming He, Xiangyu Zhang, Shaoqing Ren, and Jian Sun. 2015.
\newblock \href {http://arxiv.org/abs/1512.03385} {Deep residual learning for
  image recognition}.
\newblock \emph{CoRR}, abs/1512.03385.

\bibitem[{Hochreiter and
  Schmidhuber(1997)}]{Hochreiter:1997:LSM:1246443.1246450}
Sepp Hochreiter and J{\"u}rgen Schmidhuber. 1997.
\newblock \href {https://doi.org/10.1162/neco.1997.9.8.1735} {{Long Short-Term
  Memory}}.
\newblock \emph{Neural Comput.}, 9(8):1735--1780.

\bibitem[{Junczys-Dowmunt et~al.(2018)Junczys-Dowmunt, Heafield, Hoang,
  Grundkiewicz, and Aue}]{Junczys-Dowmunt-wnmtmarian}
Marcin Junczys-Dowmunt, Kenneth Heafield, Hieu Hoang, Roman Grundkiewicz, and
  Anthony Aue. 2018.
\newblock \href
  {https://kheafield.com/papers/edinburgh/wnmt\_marian\_paper.pdf} {Marian:
  Cost-effective high-quality neural machine translation in {C++}}.
\newblock In \emph{Proceedings of the 2nd Workshop on Neural Machine
  Translation and Generation}, Melbourne, Australia.

\bibitem[{Ott et~al.(2018)Ott, Edunov, Grangier, and Auli}]{ott2018scaling}
Myle Ott, Sergey Edunov, David Grangier, and Michael Auli. 2018.
\newblock Scaling neural machine translation.
\newblock \emph{arXiv preprint arXiv:1806.00187}.

\bibitem[{Papineni et~al.(2002)Papineni, Roukos, Ward, and
  Zhu}]{Papineni:2002:BMA:1073083.1073135}
Kishore Papineni, Salim Roukos, Todd Ward, and Wei-Jing Zhu. 2002.
\newblock \href {https://doi.org/10.3115/1073083.1073135} {Bleu: A method for
  automatic evaluation of machine translation}.
\newblock In \emph{Proceedings of the 40th Annual Meeting on Association for
  Computational Linguistics}, ACL '02, pages 311--318, Stroudsburg, PA, USA.
  Association for Computational Linguistics.

\bibitem[{Pham et~al.(2019)Pham, Nguyen, Niehues, M{\"u}ller, and
  Waibel}]{Pham2019VeryDS}
Ngoc-Quan Pham, Thai-Son Nguyen, Jan Niehues, Markus M{\"u}ller, and
  Alexander~H. Waibel. 2019.
\newblock Very deep self-attention networks for end-to-end speech recognition.
\newblock \emph{CoRR}, abs/1904.13377.

\bibitem[{Post(2018)}]{W18-6319}
Matt Post. 2018.
\newblock \href {http://aclweb.org/anthology/W18-6319} {A call for clarity in
  reporting {BLEU} scores}.
\newblock In \emph{Proceedings of the Third Conference on Machine Translation:
  Research Papers}, pages 186--191. Association for Computational Linguistics.

\bibitem[{Radford et~al.(2018)Radford, Narasimhan, Salimans, and
  Sutskever}]{radford2018improving}
Alec Radford, Karthik Narasimhan, Tim Salimans, and Ilya Sutskever. 2018.
\newblock Improving language understanding by generative pre-training.

\bibitem[{Ranzato et~al.(2016)Ranzato, Chopra, Auli, and
  Zaremba}]{DBLP:journals/corr/RanzatoCAZ15}
Marc'Aurelio Ranzato, Sumit Chopra, Michael Auli, and Wojciech Zaremba. 2016.
\newblock {Sequence Level Training with Recurrent Neural Networks}.
\newblock In \emph{{The International Conference on Learning Representations}}.

\bibitem[{Sennrich et~al.(2016)Sennrich, Haddow, and
  Birch}]{sennrich-etal-2016-neural}
Rico Sennrich, Barry Haddow, and Alexandra Birch. 2016.
\newblock \href {https://doi.org/10.18653/v1/P16-1162} {Neural machine
  translation of rare words with subword units}.
\newblock In \emph{Proceedings of the 54th Annual Meeting of the Association
  for Computational Linguistics (Volume 1: Long Papers)}, pages 1715--1725,
  Berlin, Germany. Association for Computational Linguistics.

\bibitem[{Simonyan and Zisserman(2015)}]{Simonyan15}
K.~Simonyan and A.~Zisserman. 2015.
\newblock Very deep convolutional networks for large-scale image recognition.
\newblock In \emph{International Conference on Learning Representations}.

\bibitem[{So et~al.(2019)So, Liang, and Le}]{so2019evolved}
David~R So, Chen Liang, and Quoc~V Le. 2019.
\newblock The evolved transformer.
\newblock \emph{arXiv preprint arXiv:1901.11117}.

\bibitem[{Stern et~al.(2018)Stern, Shazeer, and Uszkoreit}]{stern2018blockwise}
Mitchell Stern, Noam Shazeer, and Jakob Uszkoreit. 2018.
\newblock Blockwise parallel decoding for deep autoregressive models.
\newblock In \emph{Advances in Neural Information Processing Systems}, pages
  10086--10095.

\bibitem[{Vaswani et~al.(2018)Vaswani, Bengio, Brevdo, Chollet, Gomez, Gouws,
  Jones, Kaiser, Kalchbrenner, Parmar, Sepassi, Shazeer, and
  Uszkoreit}]{vaswani2018tensor2tensor}
Ashish Vaswani, Samy Bengio, Eugene Brevdo, Francois Chollet, Aidan Gomez,
  Stephan Gouws, Llion Jones, {\L}ukasz Kaiser, Nal Kalchbrenner, Niki Parmar,
  Ryan Sepassi, Noam Shazeer, and Jakob Uszkoreit. 2018.
\newblock \href {https://www.aclweb.org/anthology/W18-1819} {{T}ensor2{T}ensor
  for neural machine translation}.
\newblock In \emph{Proceedings of the 13th Conference of the Association for
  Machine Translation in the {A}mericas (Volume 1: Research Papers)}, pages
  193--199, Boston, MA. Association for Machine Translation in the Americas.

\bibitem[{Vaswani et~al.(2017)Vaswani, Shazeer, Parmar, Uszkoreit, Jones,
  Gomez, Kaiser, and Polosukhin}]{NIPS2017_7181}
Ashish Vaswani, Noam Shazeer, Niki Parmar, Jakob Uszkoreit, Llion Jones,
  Aidan~N Gomez, \L~ukasz Kaiser, and Illia Polosukhin. 2017.
\newblock \href
  {http://papers.nips.cc/paper/7181-attention-is-all-you-need.pdf} {Attention
  is all you need}.
\newblock In I.~Guyon, U.~V. Luxburg, S.~Bengio, H.~Wallach, R.~Fergus,
  S.~Vishwanathan, and R.~Garnett, editors, \emph{Advances in Neural
  Information Processing Systems 30}, pages 5998--6008. Curran Associates, Inc.

\bibitem[{Wang et~al.(2017)Wang, Lu, Zhou, and Liu}]{wang-etal-2017-deep}
Mingxuan Wang, Zhengdong Lu, Jie Zhou, and Qun Liu. 2017.
\newblock \href {https://doi.org/10.18653/v1/P17-1013} {Deep neural machine
  translation with linear associative unit}.
\newblock In \emph{Proceedings of the 55th Annual Meeting of the Association
  for Computational Linguistics (Volume 1: Long Papers)}, pages 136--145,
  Vancouver, Canada. Association for Computational Linguistics.

\bibitem[{Wang et~al.(2019)Wang, Li, Xiao, Zhu, Li, Wong, and
  Chao}]{wang-etal-2019-learning}
Qiang Wang, Bei Li, Tong Xiao, Jingbo Zhu, Changliang Li, Derek~F. Wong, and
  Lidia~S. Chao. 2019.
\newblock \href {https://www.aclweb.org/anthology/P19-1176} {Learning deep
  transformer models for machine translation}.
\newblock In \emph{Proceedings of the 57th Annual Meeting of the Association
  for Computational Linguistics}, pages 1810--1822, Florence, Italy.
  Association for Computational Linguistics.

\bibitem[{Wu et~al.(2019)Wu, Fan, Baevski, Dauphin, and Auli}]{wu2018pay}
Felix Wu, Angela Fan, Alexei Baevski, Yann Dauphin, and Michael Auli. 2019.
\newblock \href {https://openreview.net/forum?id=SkVhlh09tX} {Pay less
  attention with lightweight and dynamic convolutions}.
\newblock In \emph{International Conference on Learning Representations}.

\bibitem[{Wu et~al.(2016)Wu, Schuster, Chen, Le, Norouzi, Macherey, Krikun,
  Cao, Gao, Macherey, Klingner, Shah, Johnson, Liu, Łukasz Kaiser, Gouws,
  Kato, Kudo, Kazawa, Stevens, Kurian, Patil, Wang, Young, Smith, Riesa,
  Rudnick, Vinyals, Corrado, Hughes, and Dean}]{45610}
Yonghui Wu, Mike Schuster, Zhifeng Chen, Quoc~V. Le, Mohammad Norouzi, Wolfgang
  Macherey, Maxim Krikun, Yuan Cao, Qin Gao, Klaus Macherey, Jeff Klingner,
  Apurva Shah, Melvin Johnson, Xiaobing Liu, Łukasz Kaiser, Stephan Gouws,
  Yoshikiyo Kato, Taku Kudo, Hideto Kazawa, Keith Stevens, George Kurian,
  Nishant Patil, Wei Wang, Cliff Young, Jason Smith, Jason Riesa, Alex Rudnick,
  Oriol Vinyals, Greg Corrado, Macduff Hughes, and Jeffrey Dean. 2016.
\newblock \href {http://arxiv.org/abs/1609.08144} {Google's neural machine
  translation system: Bridging the gap between human and machine translation}.
\newblock \emph{CoRR}, abs/1609.08144.

\bibitem[{Zaeemzadeh et~al.(2018)Zaeemzadeh, Rahnavard, and
  Shah}]{DBLP:journals/corr/abs-1805-07477}
Alireza Zaeemzadeh, Nazanin Rahnavard, and Mubarak Shah. 2018.
\newblock \href {http://arxiv.org/abs/1805.07477} {Norm-preservation: Why
  residual networks can become extremely deep?}
\newblock \emph{CoRR}, abs/1805.07477.

\bibitem[{Zhang et~al.(2018)Zhang, Xiong, and
  Su}]{zhang-etal-2018-accelerating}
Biao Zhang, Deyi Xiong, and Jinsong Su. 2018.
\newblock \href {https://www.aclweb.org/anthology/P18-1166} {Accelerating
  neural transformer via an average attention network}.
\newblock In \emph{Proceedings of the 56th Annual Meeting of the Association
  for Computational Linguistics (Volume 1: Long Papers)}, pages 1789--1798,
  Melbourne, Australia. Association for Computational Linguistics.

\bibitem[{{Zhang} et~al.(2018){Zhang}, {Xiong}, and {Su}}]{8493282}
Biao {Zhang}, Deyi {Xiong}, and Jinsong {Su}. 2018.
\newblock \href {https://doi.org/10.1109/TPAMI.2018.2876404} {Neural machine
  translation with deep attention}.
\newblock \emph{IEEE Transactions on Pattern Analysis and Machine
  Intelligence}, pages 1--1.

\bibitem[{Zhang et~al.(2016)Zhang, Xiong, Su, Duan, and
  Zhang}]{zhang-etal-2016-variational-neural}
Biao Zhang, Deyi Xiong, Jinsong Su, Hong Duan, and Min Zhang. 2016.
\newblock \href {https://doi.org/10.18653/v1/D16-1050} {Variational neural
  machine translation}.
\newblock In \emph{Proceedings of the 2016 Conference on Empirical Methods in
  Natural Language Processing}, pages 521--530, Austin, Texas. Association for
  Computational Linguistics.

\bibitem[{Zhang et~al.(2019)Zhang, Dauphin, and Ma}]{zhang2018residual}
Hongyi Zhang, Yann~N. Dauphin, and Tengyu Ma. 2019.
\newblock \href {https://openreview.net/forum?id=H1gsz30cKX} {Fixup
  initialization: Residual learning without normalization via better
  initialization}.
\newblock In \emph{International Conference on Learning Representations}.

\bibitem[{Zhang et~al.(2018)Zhang, Su, Qin, Liu, Ji, and
  Wang}]{zhang2018asynchronous}
Xiangwen Zhang, Jinsong Su, Yue Qin, Yang Liu, Rongrong Ji, and Hongji Wang.
  2018.
\newblock Asynchronous bidirectional decoding for neural machine translation.
\newblock In \emph{Thirty-Second AAAI Conference on Artificial Intelligence}.

\bibitem[{Zhou et~al.(2016)Zhou, Cao, Wang, Li, and Xu}]{zhou-etal-2016-deep}
Jie Zhou, Ying Cao, Xuguang Wang, Peng Li, and Wei Xu. 2016.
\newblock \href {https://doi.org/10.1162/tacl_a_00105} {Deep recurrent models
  with fast-forward connections for neural machine translation}.
\newblock \emph{Transactions of the Association for Computational Linguistics},
  4:371--383.

\end{thebibliography}
\bibliographystyle{acl_natbib}

\end{document}